\documentclass[11pt,a4paper]{article}

\usepackage[utf8]{inputenc}
\usepackage[T1]{fontenc}
\usepackage{graphicx}
\usepackage{amsmath,amssymb}
\usepackage{booktabs}
\usepackage[margin=1in]{geometry}
\usepackage{caption}
\usepackage{subcaption}
\usepackage{float}
\usepackage{xcolor}
\usepackage{hyperref}
\hypersetup{
    colorlinks=true,
    linkcolor=blue!60!black,
    citecolor=blue!60!black,
    urlcolor=blue!60!black
}
\usepackage{tikz}
\usetikzlibrary{shapes.geometric,arrows.meta,positioning,fit,backgrounds,calc,decorations.pathreplacing}
\usepackage{listings}
\usepackage{fancyhdr}
\usepackage{titlesec}
\usepackage{enumitem}
\setlist[itemize]{leftmargin=*}
\setlist[enumerate]{leftmargin=*}
\usepackage{framed}
\usepackage{multirow}

\lstdefinestyle{json}{
    backgroundcolor=\color{gray!10},
    basicstyle=\small\ttfamily,
    frame=single,
    breaklines=true,
    columns=fullflexible,
    showstringspaces=false,
}

\title{Modality-Native Routing in Agent-to-Agent Networks:\\A Multimodal A2A Protocol Extension}

\author{
  Vasundra Srinivasan \\
  \textit{AI Architect, Author---Data Engineering for Multimodal AI (O'Reilly),} \\
  \textit{Stanford School of Engineering}
}

\date{April 2026}

\begin{document}
\maketitle

\begin{abstract}
Preserving multimodal signals across agent boundaries is necessary for accurate cross-modal reasoning---but it is not sufficient. We show that modality-native routing in Agent-to-Agent (A2A) networks improves task accuracy by 20 percentage points over text-bottleneck baselines, but \emph{only} when the downstream reasoning agent can exploit the richer context that native routing preserves. An ablation replacing LLM-backed reasoning with keyword matching eliminates the accuracy gap entirely (36\% vs.\ 36\%), establishing a \textbf{two-layer requirement}: protocol-level routing must be paired with capable agent-level reasoning for the benefit to materialize.

We present \textbf{MMA2A}, an architecture layer atop A2A that inspects Agent Card capability declarations to route voice, image, and text parts in their native modality. On \textsc{CrossModal-CS}, a controlled 50-task benchmark where same LLM backend, same tasks, and only the routing path varies, MMA2A achieves 52\% task completion accuracy versus 32\% for the text-bottleneck baseline (95\% bootstrap CI on $\Delta$TCA: [8, 32]\,pp; McNemar's exact $p = 0.006$). Gains concentrate on vision-dependent tasks: product defect reports improve by +38.5\,pp and visual troubleshooting by +16.7\,pp. This accuracy gain comes at a 1.8$\times$ latency cost from native multimodal processing, creating an accuracy-latency tradeoff. These results suggest that routing is a first-order design variable in multi-agent systems, as it determines the information available for downstream reasoning.
\end{abstract}

\section{Introduction}
\label{sec:intro}

When multimodal agents communicate over the Agent-to-Agent (A2A) protocol~\cite{a2a2025}, the common deployment pattern serializes all inter-agent messages to text---even though the protocol natively supports audio, image, and structured data parts. We call this the \textbf{text-bottleneck pipeline}. It discards perceptual signals (prosodic cues, spatial defect features, visual context) that downstream agents could exploit for better decisions.

Replacing text serialization with \emph{modality-native routing}---forwarding parts in their original modality when the receiving agent supports it---seems like an obvious improvement. But our experiments reveal a subtlety: \textbf{routing alone does not improve outcomes---it only changes the information available to the decision-making agent}. In an initial configuration where the final decision-making agent used keyword-matching heuristics, modality-native routing and text-bottleneck pipelines produced identical accuracy (36\% vs.\ 36\%). The richer context was delivered but never utilized. Only when we replaced the heuristic with LLM-backed reasoning did a 20 percentage point accuracy gap emerge (52\% vs.\ 32\%).

This finding motivates a \textbf{two-layer requirement} for effective multimodal agent communication: (1) a \emph{protocol layer} that preserves native modality across agent boundaries, and (2) a \emph{reasoning layer} in which agents can distinguish between high-fidelity and degraded evidence. Both are necessary; neither alone is sufficient.

We instantiate this insight in \textbf{MMA2A} (\textbf{M}ultimodal \textbf{M}odality-native \textbf{A2A}), a lightweight routing layer atop A2A that inspects Agent Card capability declarations and routes parts natively when possible. MMA2A requires no protocol modifications---it leverages existing FilePart and Agent Card features that are specified but underutilized. Our contributions:

\begin{enumerate}
    \item \textbf{MMA2A architecture}: A modality-aware router that reads Agent Card \texttt{inputModes}/\texttt{outputModes} to make routing decisions at dispatch time (\S\ref{sec:architecture}).
    \item \textbf{CrossModal-CS benchmark}: A controlled 50-task customer service benchmark requiring joint voice, image, and text reasoning (\S\ref{sec:benchmark}).
    \item \textbf{Controlled experiments with ablation}: Paired evaluation isolating routing strategy as the sole variable (same Gemini backend, same tasks, same knowledge base), plus an ablation that causally demonstrates the two-layer requirement (\S\ref{sec:experiments}).
\end{enumerate}

\section{Background and Related Work}
\label{sec:related}

\subsection{The Agent-to-Agent Protocol}

A2A~\cite{a2a2025}, contributed by Google to the Linux Foundation in 2025, defines inter-agent communication over HTTP with JSON-RPC 2.0. Agents publish \emph{Agent Cards} (JSON capability descriptors at \texttt{/.well-known/agent-card.json}) and exchange \emph{Messages} containing typed \emph{Parts}. The Part type system is central to our contribution: \textbf{TextPart} carries plain text, \textbf{FilePart} carries file content with a \texttt{mimeType} field (supporting \texttt{audio/wav}, \texttt{image/png}, etc.), and \textbf{DataPart} carries structured JSON. FilePart means the protocol already supports native multimodal payloads---a capability present in the specification but absent from practice. We reference the v0.2 specification; the v1.0 draft consolidates Part types and renames \texttt{mimeType} to \texttt{mediaType}, but our routing logic is compatible with both.

\subsection{Agent Interoperability Landscape}

Ehtesham et al.~\cite{ehtesham2025survey} survey four emerging protocols: MCP~\cite{mcp2024} (tool invocation via JSON-RPC client-server), ACP~\cite{acp2025} (RESTful multipart messaging), A2A (collaborative task execution), and ANP (decentralized agent marketplaces). They propose a phased adoption roadmap beginning with MCP for tool access, then ACP for multimodal messaging. Our work demonstrates that A2A alone is sufficient for multimodal inter-agent communication, without requiring ACP as an intermediary, by leveraging the existing FilePart and Agent Card features.

\subsection{Multi-Agent Systems with A2A}

AgentMaster~\cite{liao2025agentmaster} combines A2A and MCP within a single framework, achieving BERTScore F1 of 96.3\% and LLM-as-a-Judge G-Eval of 87.1\% on multimodal information retrieval. However, its multimodal support is text-dominant: queries are decomposed into text sub-queries, with image analysis handled by a single vision tool rather than a peer agent communicating via native image parts. It does not evaluate streaming voice input or measure the accuracy cost of modality conversion. Multi-agent orchestration surveys~\cite{orchestration2025} and gossip-based coordination proposals~\cite{gossip2025} address scalability but assume text-only payloads.

Li et al.~\cite{lacp2025} propose LACP, a telecom-inspired protocol for LLM agent communication, arguing that current protocols lack the structured session management needed for reliable agent coordination. Their focus is on protocol design rather than modality routing.

\subsection{Multimodal Foundation Models}

Recent models (GPT-4o~\cite{gpt4o2024}, Gemini~\cite{gemini2024}) natively process text, images, and audio within a single inference call. Our work is distinct: rather than relying on a monolithic multimodal model, we study \emph{distributed heterogeneous agents} that each specialize in one modality but must collaborate over A2A. This reflects enterprise deployments where different teams or vendors supply different agent capabilities, and where no single model has access to all proprietary tools and data sources. Notably, we \emph{use} a multimodal model (Gemini 2.5 Flash) as the shared backend for all agents---the point is not that the model cannot process all modalities, but that the \emph{routing decision} determines what evidence each agent receives.

Multimodal benchmarks such as MM-Bench~\cite{mmbench2024} and MMMU~\cite{mmmu2024} evaluate single-model capability on cross-modal tasks but do not address the distributed setting where modalities arrive at separate agents via a protocol. Similarly, multi-agent orchestration frameworks (AutoGen~\cite{autogen2024}, CrewAI~\cite{crewai2024}) coordinate agent workflows but assume text-only message passing. Our contribution sits at the intersection: measuring the accuracy cost of text-only routing in a \emph{multi-agent, multi-modal, protocol-mediated} system.

\section{System Architecture}
\label{sec:architecture}

\subsection{Deployment Overview}

\begin{figure}[H]
\centering
\begin{tikzpicture}[
    node distance=1.2cm and 2.5cm,
    box/.style={rectangle, draw, rounded corners, minimum width=3.2cm, minimum height=1cm, align=center, font=\small},
    agent/.style={box, fill=blue!10},
    router/.style={box, fill=orange!15},
    user/.style={box, fill=green!10},
    arr/.style={-{Stealth[length=3mm]}, thick}
]
    \node[user] (user) {User Input\\(voice + image + text)};

    \node[router, below=of user] (mar) {Modality-Aware\\Router (MAR)};

    \node[agent, below left=1.5cm and 1.5cm of mar] (voice) {Voice Agent\\$\mathcal{A}_v$\\(Gemini 2.5 Flash)};
    \node[agent, below=1.5cm of mar] (vision) {Vision Agent\\$\mathcal{A}_i$\\(Gemini 2.5 Flash)};
    \node[agent, below right=1.5cm and 1.5cm of mar] (text) {Text Agent\\$\mathcal{A}_t$\\(Gemini 2.5 Flash)};

    \node[router, below=3.5cm of mar] (orch) {Task Orchestrator};

    \draw[arr] (user) -- node[right, font=\scriptsize] {A2A Message} (mar);
    \draw[arr] (mar) -- node[left, font=\scriptsize, align=right] {audio/wav\\(native)} (voice);
    \draw[arr] (mar) -- node[right, font=\scriptsize] {image/png} (vision);
    \draw[arr] (mar) -- node[right, font=\scriptsize, align=left] {TextPart} (text);
    \draw[arr] (voice) -- (orch);
    \draw[arr] (vision) -- (orch);
    \draw[arr] (text) -- (orch);
\end{tikzpicture}
\caption{MMA2A deployment architecture. The Modality-Aware Router inspects Agent Card capabilities and forwards parts in their native modality. Annotations show the A2A Part type used for each route.}
\label{fig:architecture}
\end{figure}
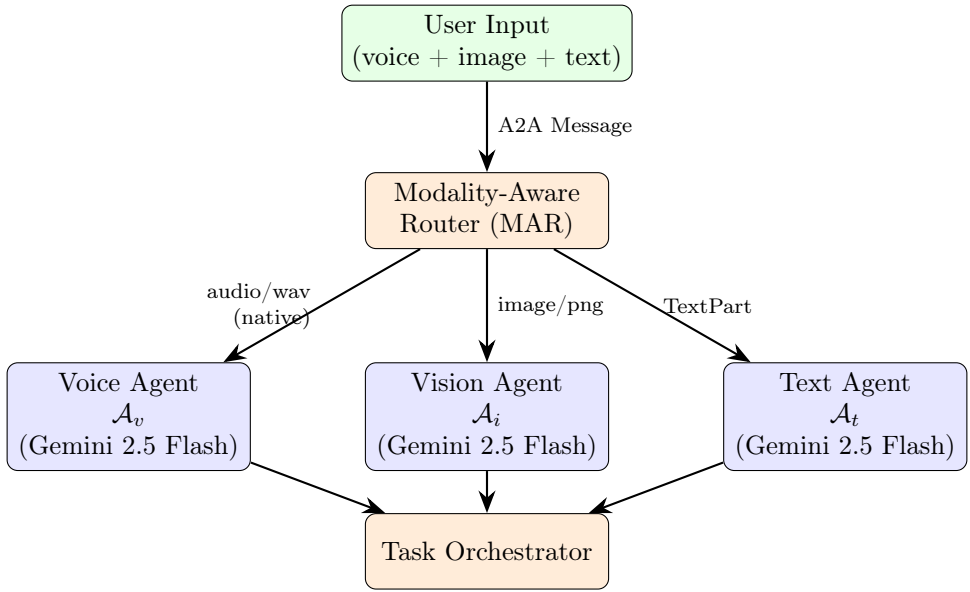

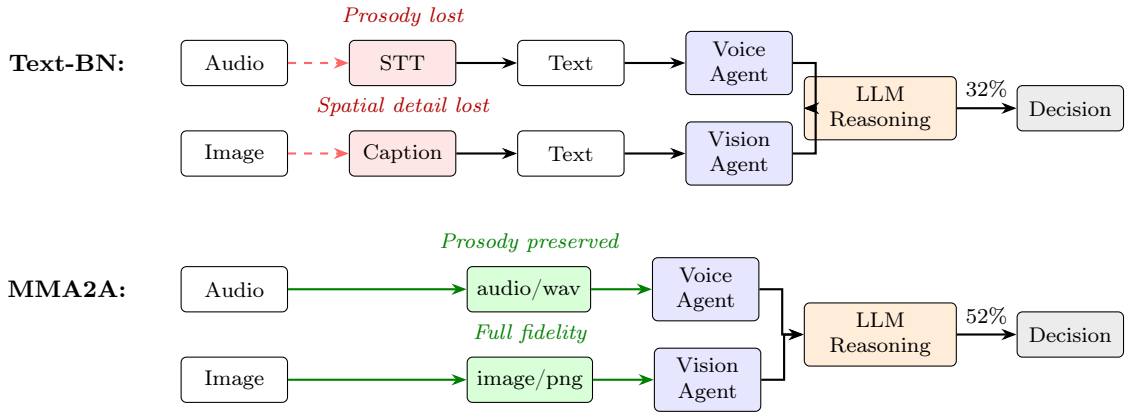
\begin{figure}[H]
\centering
\begin{tikzpicture}[
    node distance=0.6cm and 0.8cm,
    stage/.style={rectangle, draw, rounded corners=2pt, minimum width=1.4cm, minimum height=0.6cm, align=center, font=\scriptsize},
    native/.style={stage, fill=green!15},
    lossy/.style={stage, fill=red!10},
    agent/.style={stage, fill=blue!10},
    decision/.style={stage, fill=orange!15, minimum width=2cm},
    loss/.style={font=\scriptsize\itshape, text=red!70!black},
    preserve/.style={font=\scriptsize\itshape, text=green!50!black},
    arr/.style={-{Stealth[length=2mm]}, thick},
    darr/.style={-{Stealth[length=2mm]}, thick, dashed, red!60},
    garr/.style={-{Stealth[length=2mm]}, thick, green!50!black},
    lbl/.style={font=\footnotesize\bfseries},
]
    \node[lbl] at (-1.2, 0) {Text-BN:};

    \node[stage] (audio1) at (1, 0) {Audio};
    \node[lossy, right=of audio1] (stt) {STT};
    \node[stage, right=of stt] (txt1) {Text};
    \node[agent, right=of txt1] (va1) {Voice\\Agent};

    \node[stage] (img1) at (1, -1.2) {Image};
    \node[lossy, right=of img1] (cap) {Caption};
    \node[stage, right=of cap] (txt2) {Text};
    \node[agent, right=of txt2] (via1) {Vision\\Agent};

    \draw[darr] (audio1) -- (stt);
    \draw[arr] (stt) -- (txt1);
    \draw[arr] (txt1) -- (va1);
    \draw[darr] (img1) -- (cap);
    \draw[arr] (cap) -- (txt2);
    \draw[arr] (txt2) -- (via1);

    \node[loss, above=0.05cm of stt] {Prosody lost};
    \node[loss, above=0.05cm of cap] {Spatial detail lost};

    \node[lbl] at (-1.2, -3.0) {MMA2A:};

    \node[stage] (audio2) at (1, -3.0) {Audio};
    \node[native, right=2.35cm of audio2] (aud2) {audio/wav};
    \node[agent, right=of aud2] (va2) {Voice\\Agent};

    \node[stage] (img2) at (1, -4.2) {Image};
    \node[native, right=2.35cm of img2] (img2n) {image/png};
    \node[agent, right=of img2n] (via2) {Vision\\Agent};

    \draw[garr] (audio2) -- (aud2);
    \draw[garr] (aud2) -- (va2);
    \draw[garr] (img2) -- (img2n);
    \draw[garr] (img2n) -- (via2);

    \node[preserve, above=0.05cm of aud2] {Prosody preserved};
    \node[preserve, above=0.05cm of img2n] {Full fidelity};

    \node[decision] (dec1) at (9.5, -0.6) {LLM\\Reasoning};
    \node[decision] (dec2) at (9.5, -3.6) {LLM\\Reasoning};
    \node[stage, fill=gray!15] (out1) at (12, -0.6) {Decision};
    \node[stage, fill=gray!15] (out2) at (12, -3.6) {Decision};

    \draw[arr] (va1.east) -- ++(0.3,0) |- (dec1.west);
    \draw[arr] (via1.east) -- ++(0.3,0) |- (dec1.west);
    \draw[arr] (va2.east) -- ++(0.3,0) |- (dec2.west);
    \draw[arr] (via2.east) -- ++(0.3,0) |- (dec2.west);
    \draw[arr] (dec1) -- node[above, font=\scriptsize] {32\%} (out1);
    \draw[arr] (dec2) -- node[above, font=\scriptsize] {52\%} (out2);
\end{tikzpicture}
\caption{Information flow comparison. Text-BN (top) transcodes all non-text parts, losing prosodic and spatial signals. MMA2A (bottom) preserves native modality, delivering richer evidence to the shared LLM reasoning step.}
\label{fig:infoloss}
\end{figure}

Figure~\ref{fig:architecture} illustrates the MMA2A deployment architecture, and Figure~\ref{fig:infoloss} contrasts the information flow in the two pipelines we evaluate. The system consists of four components layered on standard A2A infrastructure.

\textbf{Modality-Aware Router (MAR).} A lightweight proxy ($<$ 500 lines of Python) that intercepts outgoing A2A \texttt{tasks/send} and \texttt{tasks/sendSubscribe} calls, inspects the destination agent's Agent Card for declared \texttt{inputModes} and \texttt{outputModes}, and selects the optimal part encoding. The MAR caches Agent Cards with a 60-second TTL to avoid repeated discovery round-trips. When operating in text-bottleneck mode (for baseline comparison), the MAR forces all non-text parts through transcoding regardless of the target agent's declared capabilities.

\textbf{Heterogeneous Agent Pool.} Three specialized agent types, each exposing a standard A2A Agent Card and backed by Gemini 2.5 Flash~\cite{gemini2024}:

\begin{itemize}
    \item \emph{Voice Agent} ($\mathcal{A}_v$): Processes raw audio for transcription, sentiment analysis, and urgency detection; declares \texttt{audio/wav} and \texttt{audio/webm} as input modes.
    \item \emph{Vision Agent} ($\mathcal{A}_i$): Performs visual inspection, defect detection, and image-based product identification; declares \texttt{image/png} and \texttt{image/jpeg} as input modes.
    \item \emph{Text Agent} ($\mathcal{A}_t$): Handles knowledge retrieval from a product database, warranty policy lookup, troubleshooting resolution, and final decision synthesis; operates on TextPart exclusively.
\end{itemize}

Using the same underlying model (Gemini 2.5 Flash) across all agents is a deliberate experimental design choice: it ensures that observed accuracy differences stem from the \emph{routing strategy} (what evidence reaches each agent) rather than from model capability differences.

\textbf{Task Orchestrator.} Decomposes incoming cross-modal tasks into sub-tasks, assigns them to agents via the MAR, collects results, and dispatches a final synthesis sub-task to the text agent. Independent sub-tasks (e.g., voice analysis and image analysis for the same query) are dispatched in parallel.

\subsection{Modality-Aware Routing Algorithm}

The routing decision is formalized as follows. Let $m \in \{v, i, t\}$ denote a modality (voice, image, text) and let $\text{cap}(\mathcal{A}_j)$ be the set of input modalities declared in agent $\mathcal{A}_j$'s Agent Card. Given a message part $p$ of modality $m$ destined for agent $\mathcal{A}_j$:

\begin{equation}
\text{route}(p, \mathcal{A}_j) =
\begin{cases}
p & \text{if } m \in \text{cap}(\mathcal{A}_j) \\
\text{transcode}(p, t) & \text{otherwise}
\end{cases}
\label{eq:routing}
\end{equation}

where $\text{transcode}(p, t)$ converts part $p$ to text via speech-to-text or image captioning. This rule avoids transcription when the downstream agent can natively consume the original modality, preserving information fidelity at the cost of carrying larger payloads.

\subsection{Text-Bottleneck Baseline}

The text-bottleneck (Text-BN) configuration is implemented as a single flag on the MAR that forces $\text{route}(p, \mathcal{A}_j) = \text{transcode}(p, t)$ for all non-text parts, regardless of $\text{cap}(\mathcal{A}_j)$. This simulates the dominant deployment pattern where all inter-agent communication is serialized to text. Under Text-BN, the voice and vision agents still exist and are still invoked, but they receive transcoded text descriptions rather than native audio or image data.

\subsection{A2A Protocol Compliance}

MMA2A requires \emph{no modifications} to the A2A protocol. We leverage three existing features:

\textbf{Agent Cards}: The \texttt{skills} array advertises supported MIME types in \texttt{inputModes} and \texttt{outputModes} fields, which are part of the Agent Card schema defined in the A2A specification.

\textbf{FilePart}: Audio (\texttt{audio/wav}) and image (\texttt{image/png}) data are carried inline (base64-encoded in the \texttt{data} field) for payloads under 1\,MB, or by URI reference for larger payloads. The \texttt{mimeType} field enables the receiving agent to select the appropriate decoder.

\textbf{Streaming (SSE)}: Voice interactions use \texttt{tasks/sendSubscribe} for incremental delivery, matching A2A's streaming task model.

\section{CrossModal-CS Benchmark}
\label{sec:benchmark}

We construct \textsc{CrossModal-CS}, a customer service benchmark where each task requires reasoning across at least two of three modalities. The benchmark comprises 50 tasks across four categories (Table~\ref{tab:benchmark}).

\begin{table}[H]
\centering
\caption{CrossModal-CS task categories and modality requirements.}
\label{tab:benchmark}
\begin{tabular}{@{}lcccc@{}}
\toprule
\textbf{Category} & \textbf{$n$} & \textbf{Voice} & \textbf{Image} & \textbf{Text} \\
\midrule
Product defect report   & 13 & \checkmark & \checkmark & \checkmark \\
Assembly guidance        & 12 & \checkmark &            & \checkmark \\
Visual troubleshooting  & 12 &            & \checkmark & \checkmark \\
Warranty claim          & 13 & \checkmark & \checkmark & \checkmark \\
\bottomrule
\end{tabular}
\end{table}

\textbf{Product defect report.} A customer verbally describes a product issue while sharing a photograph. The system must identify the defect from the image, cross-reference the verbal description for context (e.g., ``it started making a grinding noise after the fall''), and determine the appropriate warranty action by consulting the product knowledge base.

\textbf{Assembly guidance.} A customer asks spoken questions while following written instructions. The system must ground the spoken query in the relevant instruction step and provide actionable guidance.

\textbf{Visual troubleshooting.} A customer shares a screenshot of an error state or device condition. The system must diagnose the issue using visual evidence and match it against a troubleshooting knowledge base.

\textbf{Warranty claim.} The most complex category, requiring extraction of product identifiers from images, sentiment and urgency detection from voice, and warranty policy lookup from the text knowledge base to determine whether to approve, deny, or escalate the claim.

Each task has a ground-truth action label drawn from eight possible actions: \texttt{approve\_warranty}, \texttt{deny\_warranty}, \texttt{initiate\_replacement}, \texttt{initiate\_return}, \texttt{order\_part}, \texttt{escalate\_to\_specialist}, \texttt{provide\_instructions}, and \texttt{troubleshoot\_step}. The knowledge base contains 15 products with warranty terms and 10 troubleshooting entries.

The benchmark is deliberately small and controlled. The contribution is not a large-scale evaluation but a \emph{controlled paired experiment}: every task runs through both pipelines with identical inputs, isolating the routing strategy as the sole variable.

\section{Experiments}
\label{sec:experiments}

\subsection{Setup}

All agents run as separate A2A-compliant HTTP services on the same machine, communicating via \texttt{localhost}. Each agent is backed by Gemini 2.5 Flash via the Google AI API. The voice agent uses Gemini's multimodal input for audio transcription and analysis. The vision agent uses Gemini's image understanding for defect detection and visual diagnosis. The text agent uses Gemini for final decision synthesis, incorporating evidence collected from the other agents and the product knowledge base.

We run two configurations on all 50 tasks:

\textbf{MMA2A}: Modality-native routing per Eq.~\ref{eq:routing}. Voice parts are forwarded as \texttt{audio/wav} FileParts to the voice agent; image parts as \texttt{image/png} FileParts to the vision agent. The text agent receives synthesized text evidence from both.

\textbf{Text-BN} (Text-Bottleneck): The MAR forces all non-text parts through transcoding. Voice and image agents receive text descriptions rather than native media. This represents the dominant deployment pattern.

Both pipelines use the same Gemini 2.5 Flash model, same knowledge base, same task decomposition logic, and same final decision synthesis prompt. The only difference is whether the voice and vision agents receive native multimodal input or transcoded text.

\subsection{Metrics}

We measure three quantities:

\textbf{Task Completion Accuracy (TCA)}: The fraction of tasks where the system's recommended action exactly matches the ground-truth label (binary match).

\textbf{End-to-End Latency (E2E)}: Wall-clock time from task submission to final artifact delivery.

\textbf{Routing Profile}: The fraction of routing decisions that preserve native modality versus those that transcode to text, broken down by modality.

\subsection{Main Results}

\begin{table}[H]
\centering
\caption{Main results on CrossModal-CS (50 paired tasks). MMA2A achieves substantially higher accuracy at the cost of increased latency from native multimodal processing.}
\label{tab:results}
\begin{tabular}{@{}lcccc@{}}
\toprule
\textbf{System} & \textbf{TCA (\%)} & \textbf{Latency (s)} & \textbf{BW (KB/task)} & \textbf{Native routing (\%)} \\
\midrule
Text-BN          & 32.0 & 7.19 $\pm$ 4.46 & 329 & 50.5 \\
MMA2A (Ours)     & \textbf{52.0} & 13.04 $\pm$ 6.39 & 330 & \textbf{81.7} \\
\midrule
$\Delta$         & \textbf{+20.0\,pp} & +5.85\,s & +0.5\% & +31.2\,pp \\
\bottomrule
\end{tabular}
\end{table}

Table~\ref{tab:results} presents the main results. MMA2A improves TCA by \textbf{+20 percentage points} over Text-BN (52.0\% vs.\ 32.0\%). Of the 50 paired tasks, 11 discordant pairs favor MMA2A and 1 favors Text-BN; McNemar's exact test yields $p = 0.006$. A paired bootstrap (10{,}000 resamples) places the 95\% confidence interval on $\Delta$TCA at [8, 32]\,pp, comfortably excluding zero.

The latency difference (13.04\,s vs.\ 7.19\,s) is also statistically significant (paired $t$-test: $t = -5.76$, $p < 0.001$) and reflects the cost of native multimodal processing---the voice and vision agents perform real Gemini inference on audio and image inputs rather than operating on lossy text proxies. Bandwidth overhead is negligible (+0.5\%).

The routing telemetry confirms the experimental manipulation: MMA2A routes 81.7\% of decisions natively, while Text-BN routes only 50.5\% natively (text parts, which are native in both configurations).

\subsection{Ablation: Why Routing Alone Is Not Enough}
\label{sec:ablation}

Before analyzing the per-category results, we present an ablation that reveals the \emph{mechanism} behind MMA2A's accuracy advantage. In an initial configuration, the text agent's final decision step used a keyword-matching heuristic rather than LLM-backed reasoning. Under this configuration, both pipelines produced identical accuracy:

\begin{table}[H]
\centering
\caption{Ablation: keyword matching vs.\ LLM reasoning for the final decision step. With keyword matching, the routing strategy has no effect on accuracy.}
\label{tab:ablation}
\begin{tabular}{@{}lccc@{}}
\toprule
\textbf{Decision Method} & \textbf{Text-BN TCA} & \textbf{MMA2A TCA} & \textbf{$\Delta$} \\
\midrule
Keyword matching (heuristic) & 36\% & 36\% & 0\,pp \\
LLM reasoning (Gemini)       & 32\% & 52\% & +20\,pp \\
\bottomrule
\end{tabular}
\end{table}

The keyword matcher applies fixed rules (e.g., ``if \texttt{drop} and \texttt{damage} then \texttt{deny\_warranty}'') that produce the same output regardless of input richness---70\% of tasks yielded identical responses across both pipelines. The LLM, by contrast, can assess the \emph{quality} of the evidence, distinguishing between a detailed defect analysis from native image processing and a generic text placeholder from transcoding.

\begin{figure}[H]
\centering
\begin{tikzpicture}[
    node distance=1.8cm and 2.2cm,
    var/.style={rectangle, draw, rounded corners, minimum width=2.2cm, minimum height=0.8cm, align=center, font=\small},
    arr/.style={-{Stealth[length=3mm]}, thick},
    modarr/.style={-{Stealth[length=3mm]}, thick, dashed},
    annot/.style={font=\scriptsize\itshape, text=gray!70!black},
]
    \node[var, fill=orange!15] (routing) {Routing\\Strategy};
    \node[var, fill=blue!10, right=of routing] (fidelity) {Input\\Fidelity};
    \node[var, fill=green!15, right=of fidelity] (accuracy) {Task\\Accuracy};
    \node[var, fill=red!10, above=1.2cm of fidelity] (reasoning) {Reasoning\\Capability};

    \draw[arr] (routing) -- node[below, font=\scriptsize] {determines} (fidelity);
    \draw[arr] (fidelity) -- node[below, font=\scriptsize] {enables} (accuracy);
    \draw[modarr] (reasoning) -- node[right, font=\scriptsize, align=left] {moderates} (accuracy);

    \node[annot, below=0.3cm of routing] {MMA2A vs Text-BN};
    \node[annot, below=0.3cm of fidelity] {Native vs transcoded};
    \node[annot, below=0.3cm of accuracy] {TCA};
    \node[annot, above right=-0.15cm and 0.2cm of reasoning] {LLM vs keyword};

    \node[font=\scriptsize, text=red!60!black, below=1.0cm of fidelity, align=center] {Without reasoning: Fidelity $\to$ Accuracy path is \textbf{blocked}\\(36\% vs 36\%, $\Delta = 0$)};
\end{tikzpicture}
\caption{Causal diagram of the two-layer requirement. Routing strategy determines input fidelity, which enables accuracy gains---but only when reasoning capability (moderator) can exploit the richer evidence. The ablation confirms that blocking the moderator eliminates the effect.}
\label{fig:causal}
\end{figure}
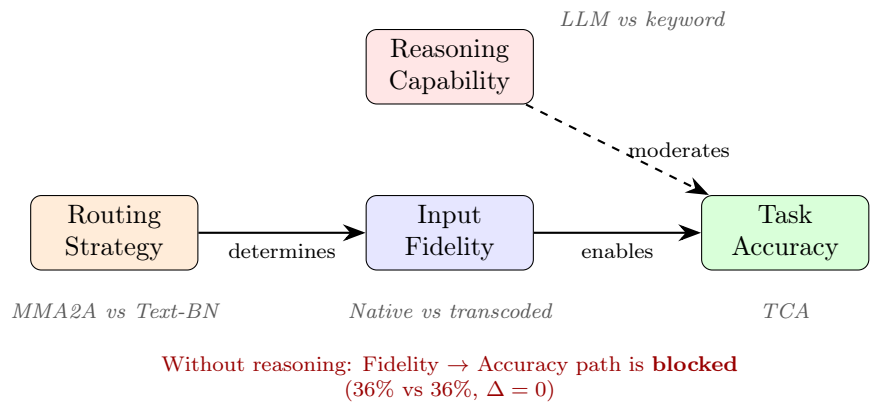

This establishes a \textbf{two-layer requirement} for modality-native routing to deliver accuracy benefits, formalized in the causal diagram of Figure~\ref{fig:causal}.

\subsection{Per-Category Analysis}

\begin{table}[H]
\centering
\caption{TCA (\%) by task category. The largest gains appear in vision-dependent categories where text transcoding loses the most signal.}
\label{tab:percategory}
\begin{tabular}{@{}lcccc@{}}
\toprule
\textbf{Category} & \textbf{$n$} & \textbf{Text-BN} & \textbf{MMA2A} & \textbf{$\Delta$} \\
\midrule
Product defect report   & 13 & 7.7  & 46.2 & \textbf{+38.5} \\
Visual troubleshooting  & 12 & 75.0 & 91.7 & +16.7 \\
Assembly guidance        & 12 & 41.7 & 58.3 & +16.6 \\
Warranty claim          & 13 & 7.7  & 15.4 & +7.7 \\
\bottomrule
\end{tabular}
\end{table}

With the two-layer mechanism established, the per-category breakdown (Table~\ref{tab:percategory}) becomes predictable: accuracy gains are largest where text transcoding discards the most decision-relevant signal.

\textbf{Product defect reports (+38.5\,pp)} show the largest improvement. These tasks depend on visual inspection---identifying a cracked screen, a bent connector, burn marks---that text captioning flattens to generic descriptions. In MMA2A mode, the vision agent processes the image natively through Gemini, producing detailed defect characterizations that the text agent's LLM reasoning can leverage. In Text-BN mode, the same vision agent receives a text description and can only echo it back, giving the decision-maker almost nothing to work with (7.7\% accuracy).

\textbf{Visual troubleshooting (+16.7\,pp, reaching 91.7\%)} confirms the ceiling: when the LLM receives high-quality visual evidence, it is highly accurate at matching error states to known troubleshooting entries.

\textbf{Assembly guidance (+16.6\,pp)} benefits because voice context carries tone and emphasis that text transcoding flattens---questions about ``this step'' versus ``the next step'' lose deictic reference in transcription. A concrete example: in task \texttt{assembly\_003}, the customer asks about crib mattress height settings via voice. The MMA2A voice agent, processing the audio natively, correctly extracts the product name and intent, and the text agent provides assembly instructions. In Text-BN mode, the transcoded text garbles the product identity---the system matches it to the wrong product entirely---and escalates to a specialist with 100\% confidence in its incorrect decision.

\textbf{Warranty claims (+7.7\,pp)} show the smallest gain. Both pipelines struggle here (15.4\% and 7.7\%). Warranty decisions depend on policy lookup and date calculations---structured data retrieval that multimodal signal alone cannot resolve. Modality-native routing helps most when the decision depends on perceptual evidence, and least when it depends on structured knowledge.

\subsection{Routing Profile}

\begin{table}[H]
\centering
\caption{Routing decisions by modality. MMA2A achieves 100\% native routing for voice and 70\% for images; Text-BN forces all non-text parts through transcoding.}
\label{tab:routing}
\begin{tabular}{@{}lcccc@{}}
\toprule
\textbf{Modality} & \multicolumn{2}{c}{\textbf{MMA2A}} & \multicolumn{2}{c}{\textbf{Text-BN}} \\
\cmidrule(lr){2-3} \cmidrule(lr){4-5}
 & Native & Transcode & Native & Transcode \\
\midrule
Voice  & 40 (100\%)  & 0 (0\%)    & 0 (0\%)    & 40 (100\%) \\
Image  & 28 (70\%)   & 12 (30\%)  & 0 (0\%)    & 40 (100\%) \\
Text   & 110 (80\%)  & 28 (20\%)  & 110 (80\%) & 28 (20\%) \\
\midrule
\textbf{Total} & 178 (82\%) & 40 (18\%) & 110 (50\%) & 108 (50\%) \\
\bottomrule
\end{tabular}
\end{table}

Table~\ref{tab:routing} shows the routing profile. MMA2A routes all voice parts natively (100\%) and most image parts natively (70\%), while Text-BN forces all non-text parts through transcoding. The 30\% image transcoding in MMA2A occurs when image parts are routed to agents that do not declare image input capabilities (e.g., the text agent).

\subsection{Latency Analysis}

\begin{table}[H]
\centering
\caption{Latency statistics (seconds) by category. MMA2A is consistently slower due to native multimodal processing, with the cost proportional to the complexity of the multimodal input.}
\label{tab:latency}
\begin{tabular}{@{}lccccc@{}}
\toprule
\textbf{Category} & \multicolumn{2}{c}{\textbf{Mean (s)}} & \multicolumn{2}{c}{\textbf{Median (s)}} \\
\cmidrule(lr){2-3} \cmidrule(lr){4-5}
 & Text-BN & MMA2A & Text-BN & MMA2A \\
\midrule
Assembly guidance        & 9.67  & 9.66  & 8.47  & 8.81 \\
Product defect report   & 3.96  & 16.55 & 3.74  & 18.19 \\
Visual troubleshooting  & 11.03 & 14.51 & 9.50  & 13.61 \\
Warranty claim          & 4.61  & 11.31 & 4.66  & 8.77 \\
\midrule
\textbf{Overall}        & 7.19  & 13.04 & 5.93  & 9.42 \\
\bottomrule
\end{tabular}
\end{table}

The latency cost of native multimodal processing is not uniform across categories (Table~\ref{tab:latency}). Product defect reports show the largest increase (3.96\,s $\to$ 16.55\,s) because native image processing requires a full Gemini inference call on the image data. Assembly guidance shows almost no latency difference (9.67\,s vs.\ 9.66\,s), suggesting that voice-only native processing adds minimal overhead in this task structure. The overall pattern is that MMA2A trades approximately 1.8$\times$ latency for a 1.6$\times$ accuracy improvement.

\subsection{Error Analysis}
\label{sec:error_analysis}

MMA2A fails on 24 of 50 tasks. To understand \emph{why}, we manually classified the 24 errors by their dominant failure mode and the architectural layer responsible (Table~\ref{tab:errors}).

\begin{table}[H]
\centering
\caption{MMA2A failure modes, classified by manual inspection of predicted-vs-expected action pairs and agent response traces. The \emph{Layer} column identifies which architectural layer is responsible for the failure.}
\label{tab:errors}
\small
\begin{tabular}{@{}lclp{4.2cm}@{}}
\toprule
\textbf{Failure Mode} & \textbf{$n$} & \textbf{Layer} & \textbf{Description} \\
\midrule
Policy lookup failure & 11 & Reasoning & Correct evidence, wrong policy match \\
Action granularity confusion & 6 & Reasoning & Semantically adjacent action selected \\
Overconfident visual grounding & 4 & Routing $\times$ Reasoning & Richer input triggers over-action \\
Insufficient context & 3 & Routing & Too little evidence reaches the agent \\
\bottomrule
\end{tabular}
\normalsize
\end{table}

The layer attribution reveals a striking asymmetry: \textbf{20 of 24 errors (83\%) originate in the reasoning layer, not in routing.} The routing layer delivers the right evidence; the reasoning layer fails to act on it correctly. This reinforces the two-layer thesis from a different angle: routing improves the \emph{inputs} to decision-making but does not resolve downstream reasoning limitations. Better routing cannot compensate for shallow knowledge retrieval or ambiguous action taxonomies.

\textbf{Policy lookup failure} is the dominant error class (11/24), concentrated entirely in warranty claims. The agent correctly identifies the product and defect but fails to retrieve the matching warranty rule from the knowledge base, defaulting to escalation. Fixing this requires richer knowledge base indexing or retrieval-augmented generation---not better routing.

\textbf{Overconfident visual grounding} (4/24) is the most interesting class because it sits at the \emph{interaction} between layers: routing delivers high-fidelity evidence, which then causes the reasoning layer to over-act. In task \texttt{defect\_006}, a peeling non-stick coating should trigger specialist escalation per company protocol, but MMA2A's vision agent produces such a detailed defect characterization that the text agent confidently initiates a replacement instead. The Text-BN baseline, receiving only a vague text description, correctly defaults to escalation. This is the only task where Text-BN outperforms MMA2A (1 of the $c = 1$ discordant pairs in McNemar's test). The implication is precise: higher-fidelity inputs can degrade performance when the optimal action depends on procedural constraints rather than perceptual evidence.

\textbf{Task trace: \texttt{defect\_001}.} To illustrate the MMA2A advantage concretely, consider a product defect report where a customer dropped a device. In MMA2A mode, the voice agent extracts the admission of a drop from the audio, the vision agent identifies physical impact damage in the image, and the text agent synthesizes both signals to correctly deny the warranty claim (physical damage exclusion). In Text-BN mode, both the voice and image parts are transcoded to text before reaching the agents. The resulting text summaries are too generic for the text agent to identify either the customer admission or the damage pattern, so it escalates to a specialist---an unnecessary and costly routing.

\section{Discussion}
\label{sec:discussion}

\subsection{The Accuracy-Latency Tradeoff}

Unlike prior work that frames multimodal optimization as a Pareto improvement (better accuracy \emph{and} lower latency), our results reveal a tradeoff. MMA2A's +20\,pp accuracy gain comes at a 1.8$\times$ latency cost. This tradeoff is intrinsic: native multimodal processing takes longer because it does \emph{more work}---the voice agent performs real speech analysis, the vision agent performs real image understanding, rather than operating on pre-transcoded text summaries.

For latency-sensitive applications (real-time customer chat), practitioners may prefer the text-bottleneck for speed, accepting lower accuracy. For accuracy-sensitive applications (warranty adjudication, safety-critical defect detection), the latency cost is acceptable. Adaptive routing that factors in task criticality is a natural extension:

\begin{equation}
\text{route}(p, \mathcal{A}_j) =
\begin{cases}
p & \text{if } m \in \text{cap}(\mathcal{A}_j) \text{ and } \text{priority}(\text{task}) \geq \theta \\
\text{transcode}(p, t) & \text{otherwise}
\end{cases}
\label{eq:adaptive}
\end{equation}

\subsection{When Does the Text-Bottleneck Suffice?}

Our warranty claim results (15.4\% vs.\ 7.7\%) suggest that for tasks dominated by structured data lookup---where the decision depends on policy terms, dates, and product identifiers rather than perceptual evidence---modality-native routing provides smaller marginal benefit. The text bottleneck may suffice when:

\begin{itemize}
    \item The task is text-only (e.g., FAQ lookup, document summarization).
    \item The perceptual content is simple enough that transcription preserves all decision-relevant information.
    \item Latency constraints preclude native multimodal processing.
\end{itemize}

MMA2A's routing rule (Eq.~\ref{eq:routing}) gracefully degrades to text-only forwarding in these cases, adding negligible overhead ($<$ 5\,ms for Agent Card capability lookup).

\subsection{Implications for A2A Ecosystem}

Our results suggest that A2A's \texttt{inputModes} and \texttt{outputModes} fields in Agent Cards are underutilized. Currently, these fields serve only as discovery metadata. We propose that A2A tooling (SDKs, orchestration frameworks) should default to modality-native routing by inspecting these fields at dispatch time, rather than assuming text serialization.

The two-layer finding (\S\ref{sec:ablation}) further implies that agent developers must design their agents to produce \emph{modality-appropriate} output: a vision agent should return structured defect assessments with severity scores, not just generic captions, so that downstream reasoning agents can distinguish high-quality from low-quality evidence. A single multimodal model could eliminate inter-agent routing entirely, but does not reflect production settings where capabilities, data access, and ownership are distributed across independently operated agents.

\subsection{Security and Privacy Considerations}

Forwarding raw audio raises privacy concerns: voice biometrics, background conversations, and ambient sounds may leak sensitive information. A2A's reliance on standard web authentication (OAuth2, API keys over TLS) mitigates inter-agent eavesdropping, but agent-level data handling policies must also be enforced. We recommend that future A2A specification revisions add a \texttt{dataRetentionPolicy} field to Agent Cards, allowing agents to declare whether they persist, cache, or immediately discard received parts.

\subsection{Routing as Information Topology}
\label{sec:topology}

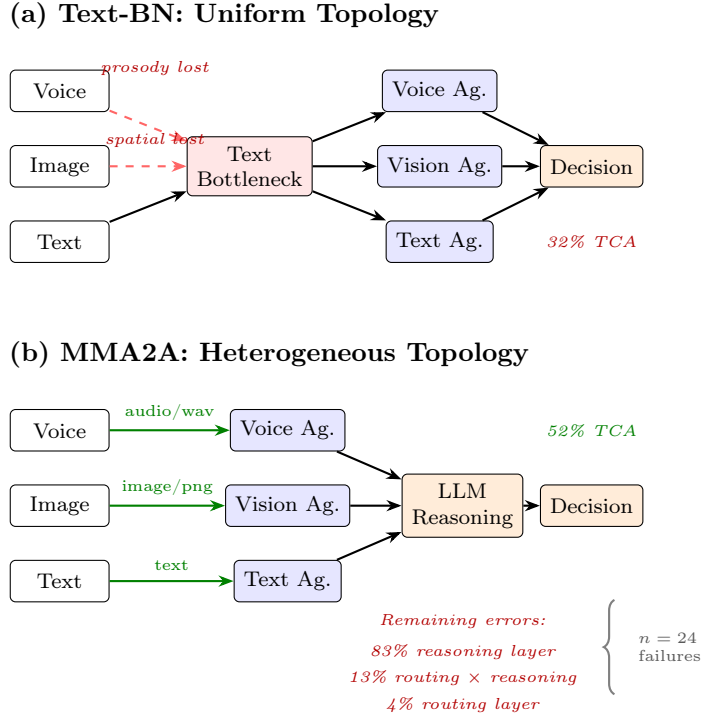
\begin{figure}[H]
\centering
\begin{tikzpicture}[
    node distance=0.7cm and 0.9cm,
    source/.style={rectangle, draw, rounded corners=2pt, minimum width=1.3cm, minimum height=0.55cm, align=center, font=\scriptsize},
    agent/.style={rectangle, draw, rounded corners=2pt, minimum width=1.3cm, minimum height=0.55cm, align=center, font=\scriptsize, fill=blue!10},
    hub/.style={rectangle, draw, rounded corners=2pt, minimum width=1.5cm, minimum height=0.55cm, align=center, font=\scriptsize, fill=red!10},
    dec/.style={rectangle, draw, rounded corners=2pt, minimum width=1.3cm, minimum height=0.55cm, align=center, font=\scriptsize, fill=orange!15},
    err/.style={font=\tiny\itshape, text=red!70!black},
    ok/.style={font=\tiny\itshape, text=green!50!black},
    arr/.style={-{Stealth[length=2mm]}, thick},
    darr/.style={-{Stealth[length=2mm]}, thick, dashed, red!60},
    garr/.style={-{Stealth[length=2mm]}, thick, green!50!black},
    lbl/.style={font=\footnotesize\bfseries},
    titl/.style={font=\small\bfseries, anchor=west},
]
    \node[titl] at (-0.3, 1.0) {(a) Text-BN: Uniform Topology};

    \node[source] (v1) at (0.5, 0) {Voice};
    \node[source] (i1) at (0.5, -1.0) {Image};
    \node[source] (t1) at (0.5, -2.0) {Text};

    \node[hub] (hub) at (3.0, -1.0) {Text\\Bottleneck};

    \node[agent] (va1) at (5.5, 0) {Voice Ag.};
    \node[agent] (ia1) at (5.5, -1.0) {Vision Ag.};
    \node[agent] (ta1) at (5.5, -2.0) {Text Ag.};

    \node[dec] (d1) at (7.5, -1.0) {Decision};

    \draw[darr] (v1) -- (hub);
    \draw[darr] (i1) -- (hub);
    \draw[arr] (t1) -- (hub);

    \draw[arr] (hub) -- (va1);
    \draw[arr] (hub) -- (ia1);
    \draw[arr] (hub) -- (ta1);

    \draw[arr] (va1) -- (d1);
    \draw[arr] (ia1) -- (d1);
    \draw[arr] (ta1) -- (d1);

    \node[err] at (1.75, 0.3) {prosody lost};
    \node[err] at (1.75, -0.65) {spatial lost};
    \node[err, align=center] at (7.5, -2.0) {32\% TCA};

    \node[titl] at (-0.3, -3.5) {(b) MMA2A: Heterogeneous Topology};

    \node[source] (v2) at (0.5, -4.5) {Voice};
    \node[source] (i2) at (0.5, -5.5) {Image};
    \node[source] (t2) at (0.5, -6.5) {Text};

    \node[agent] (va2) at (3.5, -4.5) {Voice Ag.};
    \node[agent] (ia2) at (3.5, -5.5) {Vision Ag.};
    \node[agent] (ta2) at (3.5, -6.5) {Text Ag.};

    \node[dec] (d2) at (5.8, -5.5) {LLM\\Reasoning};
    \node[dec] (d2out) at (7.5, -5.5) {Decision};

    \draw[garr] (v2) -- node[above, font=\tiny] {audio/wav} (va2);
    \draw[garr] (i2) -- node[above, font=\tiny] {image/png} (ia2);
    \draw[garr] (t2) -- node[above, font=\tiny] {text} (ta2);

    \draw[arr] (va2) -- (d2);
    \draw[arr] (ia2) -- (d2);
    \draw[arr] (ta2) -- (d2);
    \draw[arr] (d2) -- (d2out);

    \node[ok] at (7.5, -4.5) {52\% TCA};

    \node[err, align=center, anchor=north] at (5.8, -6.8) {Remaining errors:};
    \node[err, align=center, anchor=north] at (5.8, -7.2) {83\% reasoning layer};
    \node[err, align=center, anchor=north] at (5.8, -7.55) {13\% routing $\times$ reasoning};
    \node[err, align=center, anchor=north] at (5.8, -7.9) {4\% routing layer};

    \draw[decorate, decoration={brace, amplitude=4pt, mirror}, thick, gray]
        (7.8, -6.75) -- (7.8, -8.0) node[midway, right=5pt, font=\tiny, text=gray!70!black, align=left] {$n = 24$\\failures};
\end{tikzpicture}
\caption{Information topologies induced by routing strategy. (a)~Text-BN funnels all modalities through a lossy text bottleneck, imposing a uniform but degraded evidence surface on all agents. (b)~MMA2A routes each modality natively to a capable agent, creating a heterogeneous topology with higher-fidelity inputs. Error modes from Table~\ref{tab:errors} are mapped to the architectural layer responsible: the vast majority of remaining failures occur downstream of routing, in the reasoning layer.}
\label{fig:topology}
\end{figure}

The results above suggest a reframing. Routing in multi-agent systems is not merely a transport concern---it defines the \emph{information topology} of the network: which evidence is available to which agent, in what fidelity, at decision time. Figure~\ref{fig:topology} contrasts the two topologies and maps our experimental findings onto them.

The text-bottleneck imposes a \emph{uniform} topology: every agent sees the same degraded text representation regardless of the original modality (Figure~\ref{fig:topology}a). Information passes through a single lossy hub that discards prosodic cues, spatial defect features, and visual context before any agent can process them. The decision space is constrained not by the agents' capabilities but by the bottleneck's transcoding fidelity.

MMA2A restores a \emph{heterogeneous} topology (Figure~\ref{fig:topology}b) where each agent receives evidence matched to its declared capabilities, expanding the space of decisions it can make correctly. But the topology diagram also makes visible where the remaining failures concentrate: 83\% of errors occur downstream of the routing layer, in reasoning. The topology delivers the right evidence; the reasoning layer fails to exploit it.

This framing unifies our findings. The ablation (\S\ref{sec:ablation}) shows that topology alone is insufficient---agents must have reasoning capacity to exploit a richer topology. The error analysis (\S\ref{sec:error_analysis}) shows that even with the optimal topology, the reasoning layer remains the binding constraint. And the overconfident grounding cases (4/24 errors) reveal that a richer topology can \emph{narrow} the correct decision space when protocol compliance, not perceptual judgment, is required---higher-fidelity inputs are not universally beneficial.

The implication for multi-agent system design is direct: routing decisions define the information surface on which all downstream reasoning operates, and should be treated as a first-order architectural variable alongside model selection and prompt design.

\subsection{Limitations}

\textbf{Scale.} Our benchmark contains 50 tasks---sufficient for a controlled paired experiment demonstrating the effect, but not for precise effect-size estimation. A larger benchmark would enable tighter confidence intervals and per-category significance testing. We prioritized experimental control (same LLM, same tasks, same knowledge base) over scale.

\textbf{Domain.} The benchmark focuses on customer service. Generalization to other domains (healthcare imaging, manufacturing inspection, accessibility) requires domain-specific evaluation.

\textbf{Model homogeneity.} All agents use Gemini 2.5 Flash. In real deployments, agents may use heterogeneous models with different multimodal capabilities---a weaker vision model paired with a stronger text model could shift the accuracy--routing tradeoff in either direction. The interaction between model choice and routing strategy merits further study.

\textbf{Deployment conditions.} Our experiments run on a single machine with simulated A2A calls. Production deployments introduce network latency, authentication overhead, and agent availability variance that may alter the latency cost we report.

\textbf{Language.} Our experiments use English-only voice data. Multilingual evaluation would strengthen generality.

\textbf{Absolute accuracy.} MMA2A achieves 52\% TCA, which is not high in absolute terms. The benchmark includes intentionally ambiguous cases and a small knowledge base. The contribution is the \emph{relative} improvement and the per-category pattern, not the absolute level.

\section{Conclusion}
\label{sec:conclusion}

We presented MMA2A, a modality-native routing layer for the A2A protocol that enables distributed heterogeneous agents to exchange voice, image, and text parts without unnecessary transcoding. On CrossModal-CS, MMA2A improves task completion accuracy by 20 percentage points over a text-bottleneck baseline, with gains concentrated on vision-dependent tasks (+38.5\,pp for defect reports, +16.7\,pp for visual troubleshooting). This accuracy gain comes at a 1.8$\times$ latency cost from native multimodal processing.

An ablation study reveals a two-layer requirement: modality-native routing at the protocol level must be paired with LLM-backed reasoning at the agent level for the accuracy benefit to materialize. With keyword-based reasoning, the routing strategy has zero effect on accuracy.

Critically, MMA2A requires no changes to the A2A protocol---it leverages existing FilePart and Agent Card features that are specified but underutilized. Our results carry a practical message for systems builders: when deploying multi-agent systems over A2A for tasks involving cross-modal reasoning, modality-native routing is worth the latency cost.

\medskip
\noindent\textbf{Reproducibility.} All code, benchmark tasks, and configuration files are available at \url{https://github.com/vasundras/modality-native-routing-a2a-protocol}. The experiment requires a Google AI API key (Gemini 2.5 Flash) and can be reproduced on a single machine.

\bigskip
\begin{framed}
\noindent\textbf{Disclaimer.} This paper represents the author's independent research and personal views, conducted entirely outside the scope of any employment or contractual obligation. It is not sponsored by, endorsed by, affiliated with, or authorized by the author's employer, any client organization, or any technology vendor referenced herein. The author received no funding, compensation, or resources from any organization for this work. No proprietary, confidential, trade-secret, or non-public information is disclosed; all technical observations are derived solely from the author's general professional experience with publicly available protocols, open-source tools, and published specifications. All platform vendor and client organization names have been redacted to preserve confidentiality.
\end{framed}

\bibliographystyle{plain}

\end{document}